\documentclass[12pt]{iopart}
\usepackage{graphicx}
\usepackage{siunitx}
\usepackage{citesort}

\DeclareSIUnit\angstrom{\text{Å}}
\DeclareSIUnit\pixel{pixel}
\usepackage[justification=centering]{caption}
\usepackage{ragged2e}
\usepackage{hyperref}
\usepackage{cite}

\begin{document}

\title{4D-MISR: A unified model for low-dose super-resolution imaging via feature fusion}

\author{Zifei Wang$^1$, Zian Mao$^1{^,}^2{}^*$, Xiaoya He$^1$,  Xi Huang$^1$, Haoran Zhang$^1{^,}^2{}$, Chun Cheng$^1$, Shufen Chu$^1$, Tingzheng Hou$^3$, Xiaoqin Zeng$^4{^,}^5{}^*$, Yujun Xie$^1{}^*$}

\address{$^1$ Global Institute of Future Technology, Shanghai Jiao Tong University, Shanghai 200000, China}
\address{$^2$ University of Michigan-Shanghai Jiao Tong University Joint Institute, Shanghai Jiao Tong University, Shanghai 200240, China}
\address{$^3$ Institute of Materials Research, Tsinghua Shenzhen International Graduate School, Tsinghua University, Shenzhen 518055, Guangdong, China}
\address{$^4$ Materials Science and Engineering, Shanghai Jiao Tong University, Shanghai 200000, China}
\address{$^5$ National Engineering Research Center of Light Alloy Net Forming and State Key Laboratory of Metal Matrix Composites, Shanghai Jiao Tong University, Shanghai 200240, China}

\begin{abstract}
While electron microscope offers crucial atomic-resolution insights into structure–property relationships, radiation damage severely limits its use on beam-sensitive materials like proteins and 2D materials. To overcome this challenge, we push beyond the electron dose limits of conventional method by adapting principles from multi-image super-resolution (MISR) that had been widely used in remote sensing. Our method fuses multiple low-resolution, sub-pixel-shifted views and enhances this reconstruction with a convolutional neural network (CNN) that integrates features from synthetic, multi-angle observations. We developed a dual-path, attention-guided network for 4D-STEM that achieves atomic-scale super-resolution from ultra-low-dose data. This provides robust atomic-scale visualization across amorphous, semi-crystalline, and crystalline beam-sensitive specimens. Systematic evaluations on representative materials demonstrate the comparable spatial resolution to conventional ptychography under ultra low‑dose conditions. Our work one-stem expands the capabilities of 4D-STEM, offering a new and generalizable method for the structural analysis of any radiation-vulnerable material.
\end{abstract}

%

\section{Introduction}
Electron microscope, while an essential high-resolution analytical tool, faces a significant challenge in imaging electron-beam-sensitive materials like polymers, biological proteins, and two-dimensional (2D) systems \cite{lv2022low,liu2020bulk,zhou2021electron,ghosh2019electron,kuccukouglu2024low,li2025atomically,zhou2020low}. This limitation stems from electron-beam-induced damage, both chemical (e.g., radiolysis in metal-organic frameworks, MOFs, and covalent organic frameworks, COFs) and physical (e.g., atomic displacements, localized heating in polymers) \cite{patra2024recent,gryzinski1965classical,xu2023electron,jiang2020situ}. Such damages restrict atomic-level structural understanding and engineered properties, thereby narrowing their applications in electronics, catalysis, and energy storage. Although cryogenic temperatures can mitigate some damage, this approach requires specialized equipment and meticulous sample preparation \cite{zhan2024atomic,xie2025pi,xie2023spatially}.

Paradoxically, achieving high-resolution imaging demands a sufficient electron dose for an adequate signal-to-noise ratio (SNR), but higher doses inevitably exacerbate beam damage \cite{yuan2022high}. While strategies like low-dose imaging, dose fractionation, and reduced accelerating voltages can suppress knock-on damage, they often degrade resolution due to heightened chromatic aberrations and require complex post-processing\cite{radic2022treating,quigley2022cost}. This intrinsic trade-off between radiation damage and resolution severely constrains imaging capabilities for beam-sensitive materials, underscoring an urgent need for new methodologies.

The convergence of direct electron detection with four-dimensional scanning transmission electron microscopy (4D-STEM) has catalyzed transformative advances in beam-sensitive materials characterization \cite{levin2021direct,bustillo20214d,panova2019diffraction}. This multimodal platform generates rich, multidimensional datasets for nanostructural information, including crystallographic orientation, domain topology, and local heterogeneity, through optimized signal acquisition. Such highly controllable operation enables atomic-scale visualization of both soft matter (e.g., supramolecular assemblies, biomacromolecules) and inorganic architectures (e.g., MOFs, hybrid perovskites) while preserving structural integrity.

Building on these developments, coherent electron ptychography, particularly Rodenburg's iterative engine (PIE), has emerged as a powerful 4D-STEM complement \cite{hoppe1969beugung1,hoppe1969beugung2,hoppe1969beugung3}. This phase-retrieval algorithm reconstructs complex wavefunctions from overlapping diffraction patterns, enabling dose-efficient, aberration-corrected imaging without strict periodicity. However, ptychography's broader application faces two interdependent constraints: (1) irreversible beam damage exacerbated by extended acquisition times \cite{xu2025unravelling}, and (2) intrinsic SNR limitations under low-dose regimes \cite{maigne2018low}. This fundamental trade-off arises from Poissonian shot noise dominance, in which higher spatial sampling reduces electrons per diffraction pattern, causing exponential SNR degradation. Early ptychography required $1 \times 10^5 \mathrm{e}^-/\text{\AA}^2$ to $1 \times 10^6 \mathrm{e}^-/\text{\AA}^2$ for atomic resolution in radiation-tolerant samples \cite{song2019atomic}, precluding studies on sensitive systems. Even with detector and algorithmic advancements, SNR remains a critical bottleneck: below $1 \times 10^4 \mathrm{e}^-/\text{\AA}^2$, markedly deteriorates, leading to noise amplification, contrast inversion, and potential nonphysical solutions in reconstructions \cite{li2025atomically}. This effectively caps achievable resolution under low-dose conditions, as exemplified by a cryo-ptychographic study of rotavirus at $5.7  \mathrm  {e}^-/ \text{\AA}^2$ plateauing at ~3 nm resolution \cite{zhou2020low}. To be specific, this SNR collapse manifests as critically attenuated low-angle scattering signals that is essential for phase contrast and complete obliteration of high-angle components encoding sub-nanometer features. Consequently, uncorrected reconstructions exhibit pronounced noise amplification and contrast inversion unless specialized denoising protocols are applied \cite{yamada2024ptychographic}. Contradictorily, while low-dose conditions suppress uncorrelated noise through averaging effects, they simultaneously amplify structured artifacts due to ill-posed Fourier inversion. When SNR drops below critical thresholds, reconstructions often converge to nonphysical solutions, further complicating data interpretation. Therefore, the compromised SNR ultimately limits the fidelity of reconstructed phase images. Under high-dose conditions ($> 1 \times 10^4 \mathrm {e}^-/ \text{\AA}^2$), ptychography achieves full spatial frequency recovery with continuous phase gradients, enabling true atomic resolution. However, low-dose regimes ($< 1 \times 10^4 \mathrm {e}^-/ \text{\AA}^2$) induce severe high-frequency signal attenuation, effectively capping the achievable resolution. 

Drawing inspiration from multi-image super-resolution (MISR), which is a cornerstone technique in remote sensing and satellite imaging that synthesizes high-resolution images from multiple low-resolution observations of the same scene, we adapt its core principles to transcend these resolution limits in beam-sensitive material characterization. Traditional MISR exploits angular diversity across overlapping, sub-pixel-shifted low-resolution frames to recover latent spatial details. In this work, we redefine this paradigm for 4D-STEM by leveraging probe-position diversity and temporal redundancy across successive diffraction frames, effectively translating MISR’s "multi-angle" philosophy into the reciprocal-space domain. Here, we develop 4D-MISR, a unified deep learning-augmented reconstruction pipeline that integrates real-space constraints with a dual-stage neural architecture. This framework first employs a feature fusion network to aggregate spatially correlated information from overlapping 4D-STEM scan positions, mimicking MISR's cross-frame detail extraction. Subsequent stages apply iterative phase retrieval constrained by a physics-informed denoiser, enabling robust inversion of low-dose datasets at doses as low as 200 $\mathrm {e}^-/ \text{\AA}^2$. We validated 4D-MISR across model systems spanning crystalline, semi-crystalline, and amorphous materials. Compared to conventional ptychography, our method not only achieves resolution enhancement but also delivers a $>$ 3× improvement in contrast-to-noise ratio (CNR) under low-dose conditions. These results pave a new way for high-resolution imaging of beam-sensitive materials.

\section{Methods}

\subsection{Deep Learning}
\subsubsection{Model Architecture}\label{model}

\begin{figure}[htbp]
    \centering
    \includegraphics[width=0.95\textwidth]{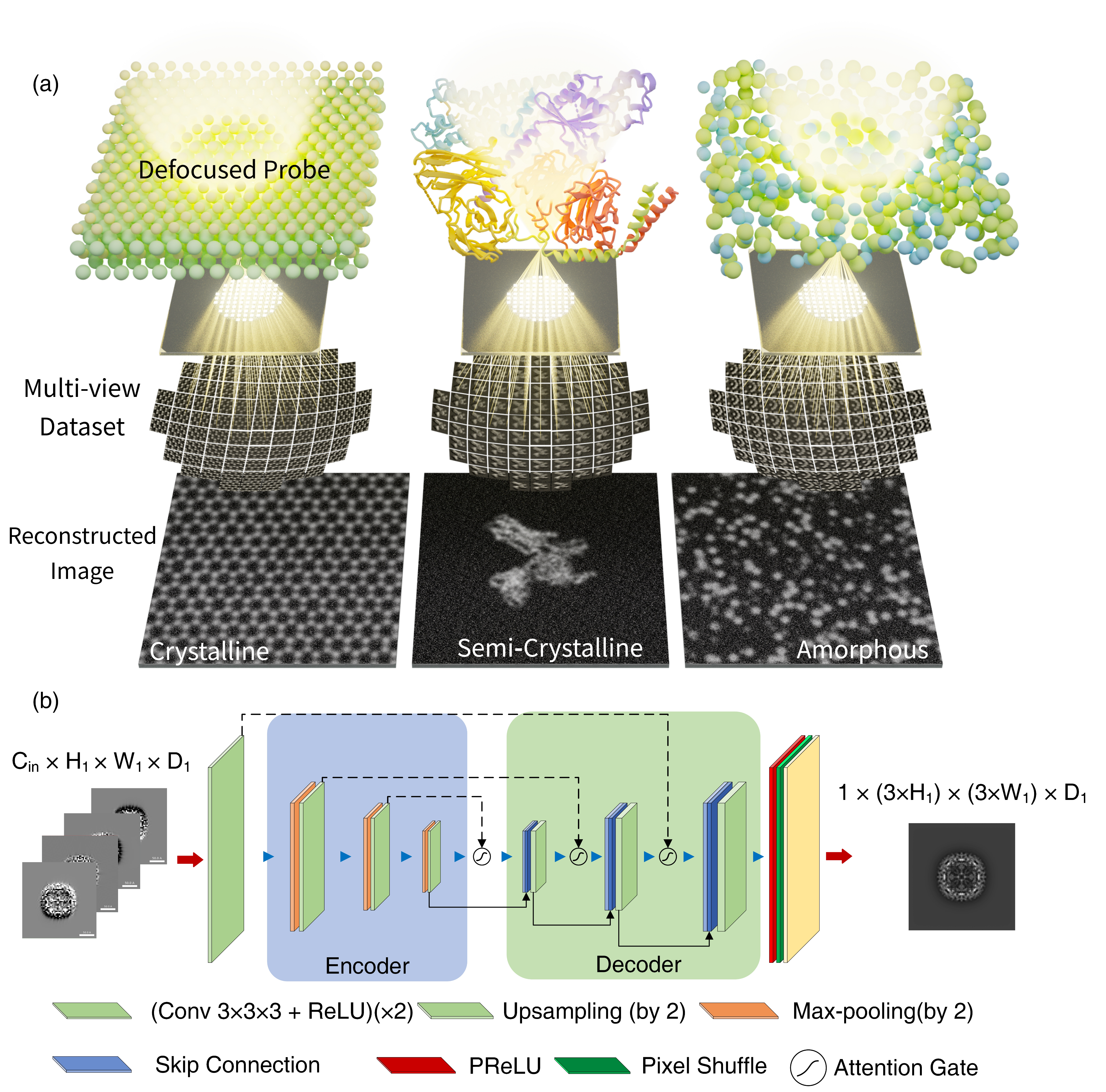} 
    \caption[Overview of proposed workflow]{
    \justifying
    (a) Multi-view electron optical configurations for 4D-STEM with defocus. In three representative systems ranging from crystalline, semi-crystalline to amorphous solids, the electron beam generates a series of images at different views, which are subsequently used for reconstruction.(b) The encoder encodes all the VBFs into embedding vectors. Attention gates determine the weights at every pixel of each VBF. The decoder uses the extracted features to generate a super-resolution image.}
    \label{fig:model}
\end{figure}

The 4D-MISR employs a two-stage architecture, comprising an Attention U-Net~\cite{oktay2018attention} followed by a sub-pixel convolutional upsampling module (Figure~\ref{fig:model}(b)). In the first stage, the input tensor with shape $(N, M, H, W)$ is processed by an encoder--decoder network equipped with attention gates, allowing the model to focus on salient spatial features while suppressing irrelevant ones. In the second stage, the output is passed through a parametric ReLU (PReLU) activation, upsampled using the PixelShuffle operation~\cite{shi2016real} and then passed through a $1 \times 1$ convolutional layer to generate a high-resolution image with scale factor $r$.


The encoder consists of five convolutional blocks, each applying two successive $3\times3$ convolutions (padding = 1), followed by batch normalization and ReLU activation. Channel depth increases progressively from 64 to 1024, with max-pooling (kernel size = 2, stride = 2) down-sampling the spatial resolution. Formally, the $i$-th encoder block computes:
\begin{equation}
X_i = \mathrm{ReLU}\bigl(\mathrm{BN}\bigl(\mathrm{Conv}_{3\times3}(X_{i-1})\bigr)\bigr),
\end{equation}
repeated twice per block, where $X_0$ is the input tensor.

The decoder mirrors the encoder with nearest-neighbor interpolation for upsampling, followed by a $3\times3$ convolution, batch normalization, and ReLU. Before skip connections are concatenated, both decoder and encoder features are passed through an attention gate. Each gate receives a gating signal $g$ from the decoder and an input $x$ from the encoder. These are projected into an intermediate space of dimensionality $F_\mathrm{int}$ via $1\times1$ convolutions and batch normalization. The attention map is computed as:
\begin{equation}
x' = x \odot \psi\bigl(\mathrm{ReLU}(W_g(g) + W_x(x))\bigr),
\end{equation}
where $\odot$ is element-wise multiplication, and $\psi$ denotes a sigmoid activation.

After decoding, the network outputs a feature tensor of shape $(r^2, H, W)$, where $r = 3$ is the super-resolution scale. This tensor is passed through a PReLU activation to introduce adaptability and prevent the dying ReLU issue~\cite{lu2019dying}. The PixelShuffle then rearranges channels into a higher-resolution image of shape $(1, rH, rW)$ via a depth-to-space transform.  A subsequent $1\times1$ convolution is applied after the sub-pixel rearrangement to act as a simple filtering mechanism for the high-frequency "checkerboard" artifacts introduced by the PixelShuffle operation. This convolution effectively mitigates the grid-like visual patterns, thereby significantly reducing the perceptible checkerboard effect.

This hybrid architecture integrates attention mechanisms to enhance sensitivity to high-frequency features and uses sub-pixel convolution for efficient and artifact-free upsampling, achieving a strong balance between computational efficiency and image reconstruction quality.

\subsubsection{Model Training}
In our training dataset, each low-dose 4D-STEM measurement is paired with its corresponding high-resolution image reconstructed via an ideal infinite-dose ptychographic algorithm. Specifically, the network inputs are the low-dose VBF images, while the targets are the high-fidelity ptychographic reconstructions that would be obtained under an infinite-dose assumption. All the datasets are generated with infinite dose at first. After that when training neural network, then every epoch we will apply different electron dose per probe and Gaussian noise to the 4D-STEM dataset rather than input 4D-STEM dataset with low electron dose. With this training strategy, on‐the‐fly noise injection affords precise control over instrument‐specific noise statistics, generates unlimited variants per clean VBF to enhance data diversity, prevents overfitting to fixed dataset artifacts, and integrates seamlessly with augmentation pipelines—thereby yielding models that are both robust and readily generalizable to real‐world EM applications. The 4D-MISR are implemented in \texttt{PyTorch} \cite{imambi2021pytorch}.

\subsubsection{Loss Function}
To effectively guide the training of our image super-resolution network, we designed a composite loss function that integrates pixel-wise reconstruction accuracy, perceptual similarity, and structural consistency. Here, the step size denotes the real‑space sampling interval of the 4D‑STEM probe, i.e. the distance between adjacent scan positions. The total loss function is formulated as:

\begin{equation}
\mathcal{L}_{\text{total}} = \mathcal{L}_{\text{pixel}} + \lambda \cdot \mathcal{L}_{\text{perceptual}} + \mathcal{L}_{\text{SSIM}}, 
\quad \text{for step size } < 1~\text{\AA}
\end{equation}

\begin{equation}
\mathcal{L}_{\text{total}} = \mathcal{L}_{\text{pixel}} + \mathcal{L}_{\text{SSIM}}, 
\quad \text{for step size } > 1~\text{\AA}
\end{equation}

where $\mathcal{L}_{\text{pixel}}$ is the standard $L_1$ loss function that computes the mean absolute error (MAE) between the predicted super-resolved image and the ground truth:

\begin{equation}
\mathcal{L}_{\text{pixel}} = \frac{1}{N} \sum_{i=1}^{N} |\hat{I}_i - I_i|.
\end{equation}

The perceptual loss $\mathcal{L}_{\text{perceptual}}$ is defined as the mean squared error (MSE) between the feature maps extracted from a pre-trained VGG19 network:

\begin{equation}
\mathcal{L}_{\text{perceptual}} = \frac{1}{C H W} \left\| \phi(\hat{I}) - \phi(I) \right\|_2^2,
\end{equation}

where $\phi(\cdot)$ denotes the activation maps obtained from the first 19 layers of the VGG19 network, and $C$, $H$, and $W$ represent the dimensions of the feature maps. Since the VGG19 network is trained on RGB images, the grayscale input images are duplicated across three channels before feeding into the network.

The structural similarity component $\mathcal{L}_{\text{SSIM}}$ is defined using the multi-scale structural similarity (MS-SSIM) index, which captures image structural information at different resolutions:

\begin{equation}
\mathcal{L}_{\text{SSIM}} = 1 - \text{MS-SSIM}(\hat{I}, I).
\end{equation}

The balancing factor $\lambda$ is empirically set to $0.006$, as suggested in prior works such as SRGAN \cite{ledig2017photo}, to maintain a reasonable trade-off between low-level accuracy and high-level perceptual quality.

This composite loss function ensures that the network not only minimizes pixel-wise errors but also generates images that are perceptually closer to the target and structurally more consistent, leading to improved performance in both objective and subjective evaluation metrics.

\subsection{Principle and General Workflow}

The key of imaging beam-sensitive materials lies in maximizing the information extracted from low-dose 4D-STEM datasets, thereby enabling high-resolution reconstruction while minimizing dose-induced artifacts. In 4D-STEM, a bright-field image is synthesized by integrating the intensity within a circular region surrounding the unscattered central beam in each diffraction pattern and assigning this integrated value to the corresponding probe position in real space \cite{ophus2019four}. By the principle of reciprocity, selecting individual pixels within this central region to construct virtual images corresponds to illuminating the sample with tilted plane waves. Thus, a single 4D-STEM scan inherently encodes a multitude of low-resolution virtual bright-field (VBF) views, each representing a unique angular perspective of the specimen(Figure~\ref{fig:model}(a)).

This perspective directly parallels the concept of multi-view geometry, where capturing a scene from different viewpoints or time frames enables recovery of high-frequency structural details \cite{luo2024large}. While extensively leveraged in remote sensing and computer vision \cite{an2022tr}, such angular diversity remains largely untapped in electron microscopy. Motivated by this analogy, we reinterpret the 4D-STEM dataset as a structured collection of angularly diverse, spatially redundant VBF images, and harness this redundancy to achieve super-resolved reconstructions, particularly under stringent dose constraints.

\begin{figure}[htbp]
    \centering
    \includegraphics[width=0.95\textwidth]{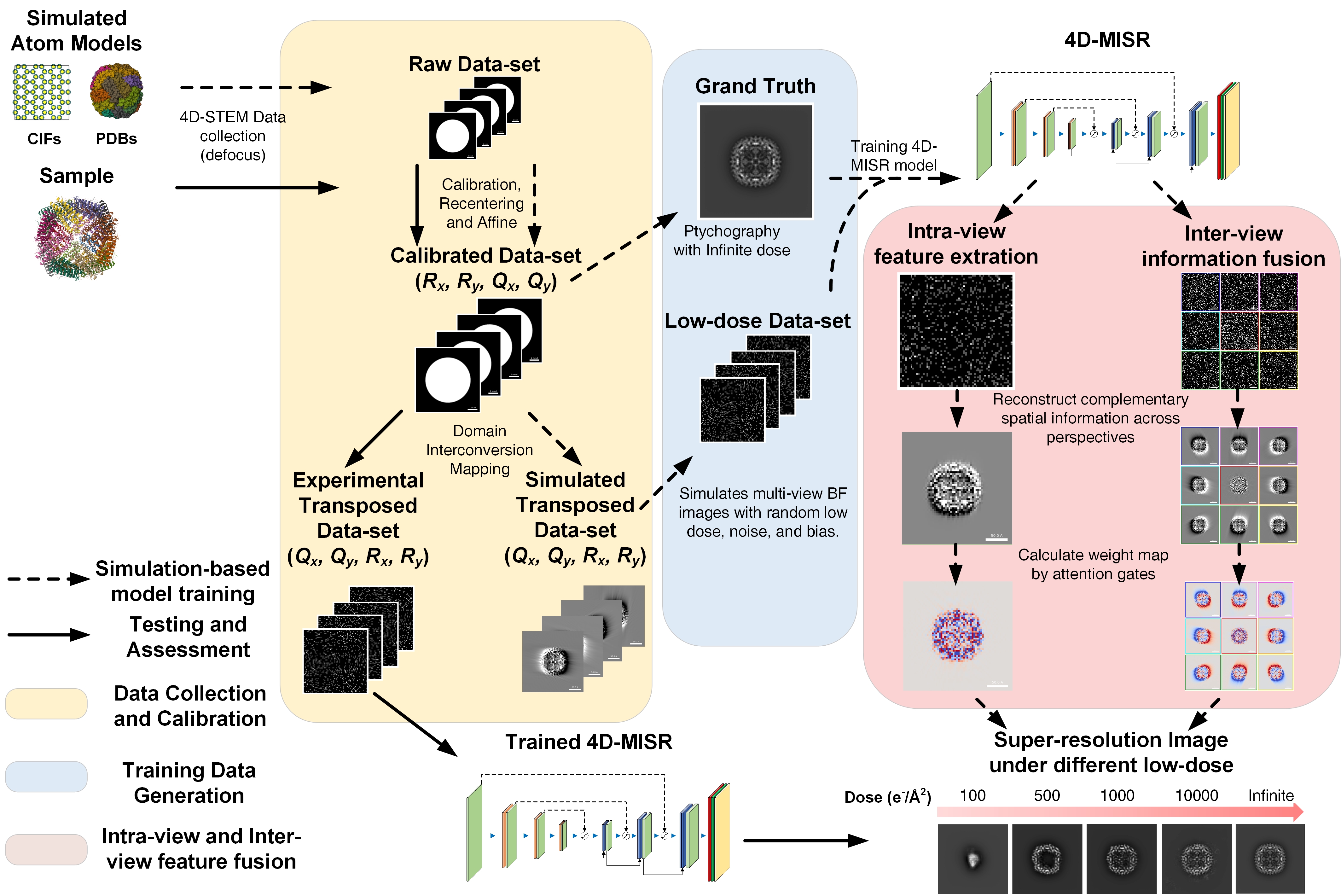} 
    \caption{\textbf{Overview of proposed workflow.} 
    \justifying
    The pipeline integrates simulation-based training data generation from atomic models, experimental data calibration with domain interconversion mapping, and a 4D-MISR deep learning model that performs intra-view feature extraction and inter-view information fusion to reconstruct high-fidelity super-resolution images across a range of electron dose conditions.}
    \label{fig:general}
\end{figure}

Figure~\ref{fig:general} illustrates a comprehensive pipeline for generating super-resolution images from 4D-STEM data using 4D-MISR. The process begins with simulated atomic models derived from Crystallographic Information File (CIF) sourced from the Materials Project~\cite{jainmaterials} and Program Database File (PDB) obtained from the Protein Data Bank~\cite{berman2000protein}, which are used to generate synthetic samples. 4D-STEM data is collected under varying defocus conditions and undergoes calibration, recentering, and affine transformation to produce a calibrated dataset $(R_{x}, R_{y}, Q_{x}, Q_{y})$. This dataset is then processed through domain interconversion mapping to generate transposed datasets $(Q_{x}, Q_{y}, R_{x}, R_{y})$. Ground truth images are obtained from infinite-dose ptychographic reconstructions, which distill the maximum achievable structural information. Corresponding low-dose datasets are simulated by introducing varying levels of noise, bias, and electron dose, mimicking practical acquisition conditions. These paired datasets are used to train the 4D-MISR network. 

The 4D-MISR network is designed to handle arbitrary collections of low-resolution views and produce a single high-resolution image by jointly leveraging spatial and angular features. The encoder extracts fine-grained spatial features from each view independently, applying attention gates to focus on locally informative details. Concurrently, an inter-view branch fuses features across views to capture global angular correlations and parallax. Two complementary update pathways refine features: one uses angular context to enhance spatial features, and the other uses spatial cues to refine angular embeddings. Multi-scale attention gates dynamically weigh these contributions, ensuring the final output preserves both high spatial detail and coherent multi-view consistency, demonstrating progressively improved image quality and validating the model's robustness and effectiveness in low-dose imaging scenarios.

\subsection{Atomic Model Generation and Multi-slice Simulation}
The simulated dataset was generated using the open-source \texttt{abTEM} module for multi-slice propagation~\cite{madsen2021abtem}. 4D-STEM simulations were carried out with a \SI{300}{\kilo\electronvolt} electron probe rastered over a $64 \times 64$ grid of probe positions at a sampling interval of \SI{4}{\angstrom\per\pixel}. The probe convergence semi-angle was set to \SI{10}{\milli\radian}, and a defocus of \SI{1500}{\angstrom} was applied. High-resolution ground-truth training images were obtained via ptychographic reconstruction using the \texttt{py4DSTEM} software package~\cite{savitzky2021py4dstem}. A complete summary of all 4D-STEM simulation parameters is provided in Supplementary Table~1.

\section{Results}
\subsection{Low-dose Imaging Across Structural Orders}
\begin{figure}[htbp]
    \centering
    \includegraphics[width=0.95\textwidth]{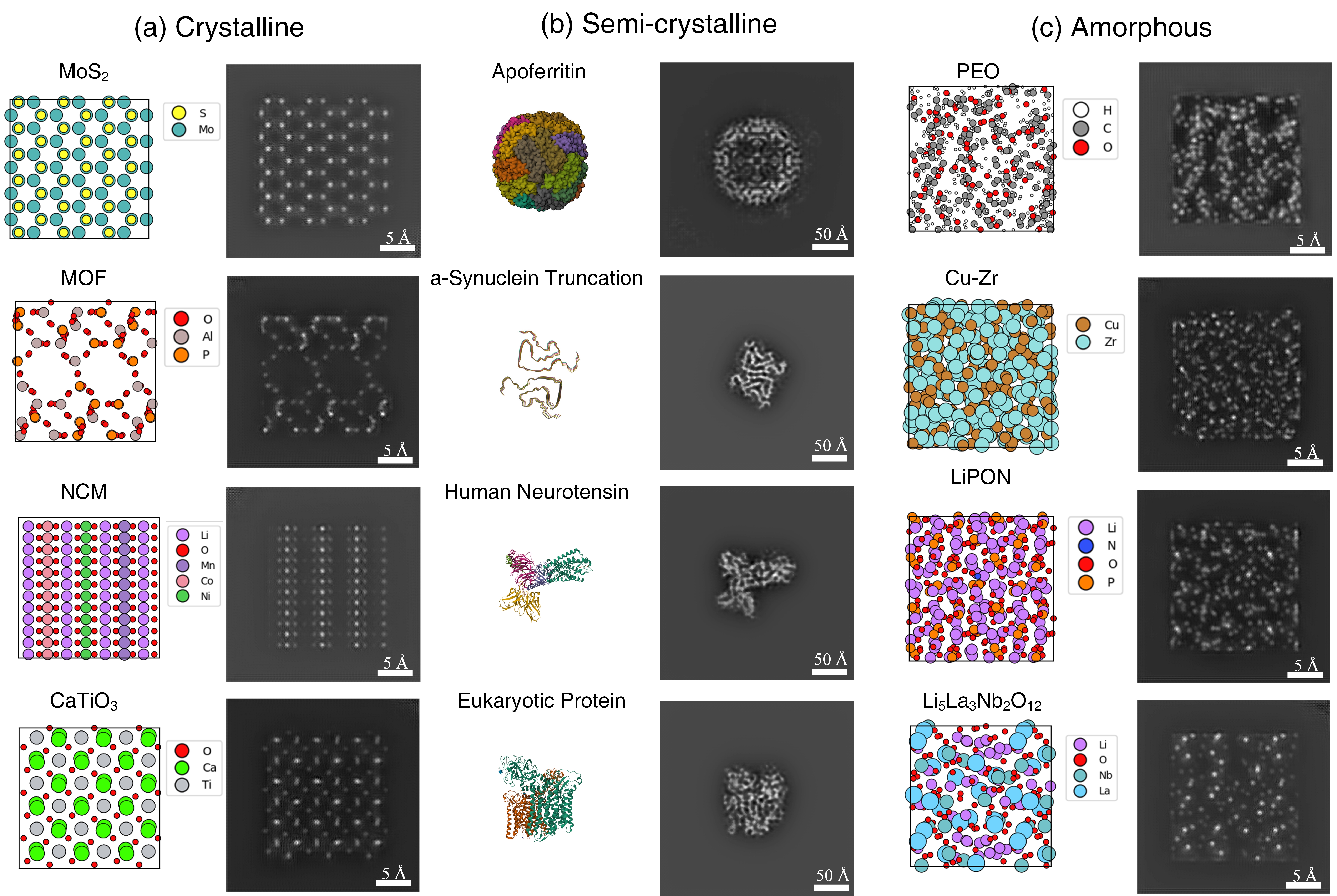} 
    \caption{\textbf{4D-MISR reconstruction results under low electron dose for materials with different structural orders.}
    \justifying
    (a) Crystalline materials (MoS$_{2}$, MOF, NCM, CaTiO$_{3}$) show clear lattice periodicity. (b) Semi-crystalline biomolecules (apoferritin, $\alpha$-synuclein truncation, neurotensin, eukaryotic protein) retain overall morphology. (c) Amorphous materials (PEO, Cu-Zr, LiPON, Li$_{5}$La$_{3}$Nb$_{2}$O$_{12}$) display disordered atomic arrangements. Left: structural models; Right: 4D-MISR results. Scale bars: \SI{5}{\angstrom} (a, c), \SI{50}{\angstrom} (b).}
    \label{fig:3}
\end{figure}

To validate the performance of our 4D-MISR method under low-dose imaging conditions, we applied it to a range of representative samples encompassing crystalline, semi-crystalline, and amorphous structures, as illustrated in Figure~\ref{fig:3}. For crystalline materials (Fig.~\ref{fig:3}a), the reconstructed images clearly resolve periodic atomic arrangements, highlighting the method’s ability to preserve fine structural details. Molybdenum disulfide (MoS$_2$) features a layered hexagonal structure composed of molybdenum atoms sandwiched between layers of sulfur atoms. The metal-organic framework (MOF) sample, consisting of aluminum, phosphorus, and oxygen atoms, exhibits a porous and periodic network typical of such frameworks. nickel cobalt manganese oxide (NCM), a layered oxide cathode material widely used in lithium-ion batteries, displays an ordered arrangement of Li, Ni, Co, and Mn ions within an oxygen matrix. Calcium titanate (CaTiO$_3$), a perovskite oxide, exhibits a cubic lattice with alternating Ca and Ti cations surrounded by oxygen anions. In all these cases, the 4D-MISR reconstruction accurately resolves periodic atomic lattices and sharp contrast, demonstrating the method's high fidelity in capturing crystallographic details.

In the case of semi-crystalline biomolecules (Fig.~\ref{fig:3}b), 4D-MISR effectively recovers overall molecular morphology despite the presence of structural heterogeneity. Apoferritin\cite{kuccukouglu2024low}, a spherical iron-storage protein, serves as a standard for cryo-EM validation due to its well-defined cage-like architecture. $\alpha$-synuclein truncation\cite{ni2019structural}, associated with neurodegenerative disorders, adopts flexible conformations and exhibits less regular structural features. Neurotensin\cite{kato2019conformational} and the eukaryotic protein\cite{bai2019structure} samples present tertiary structures with varying degrees of order. In all cases, 4D-MISR successfully recovers the global morphology and domain-level organization, indicating its robustness against local structural disorder and conformational variability.

For amorphous systems (Fig.~\ref{fig:3}c), the reconstructions exhibit no long-range order, faithfully reproducing their disordered nature typically observed using conventional method. These results collectively demonstrate the broad applicability and robustness of 4D-MISR across diverse material systems, even under stringent low-dose conditions. Polyethylene oxide (PEO) \cite{zheng2024ab} is a biocompatible, water‐soluble polymer characterized by its high flexibility and unique rheological behavior. Cu-Zr exhibits a dense amorphous network formed by rapidly quenched copper and zirconium atoms. LiPON \cite{seth2025investigating}is a solid electrolyte characterized by a disordered arrangement of Li, P, O, and N atoms. Li$_5$La$_3$Nb$_2$O$_{12}$, although Often partially crystalline, in this context presents a glassy nature. The reconstructed images from 4D-MISR preserve the lack of periodicity and yield realistic representations of the disordered atomic configurations, emphasizing the method's ability to handle structural randomness without introducing artifacts.

These comprehensive results demonstrate that 4D-MISR is capable of accurately reconstructing high-quality structural information from low-dose 4D-STEM datasets across a broad spectrum of material types, highlighting its potential for generalized, dose-efficient imaging of heterogeneous systems.

\subsection{Super-Resolution under Low Electron Dose}
\begin{figure}[htbp]
    \centering
    \includegraphics[width=0.95\textwidth]{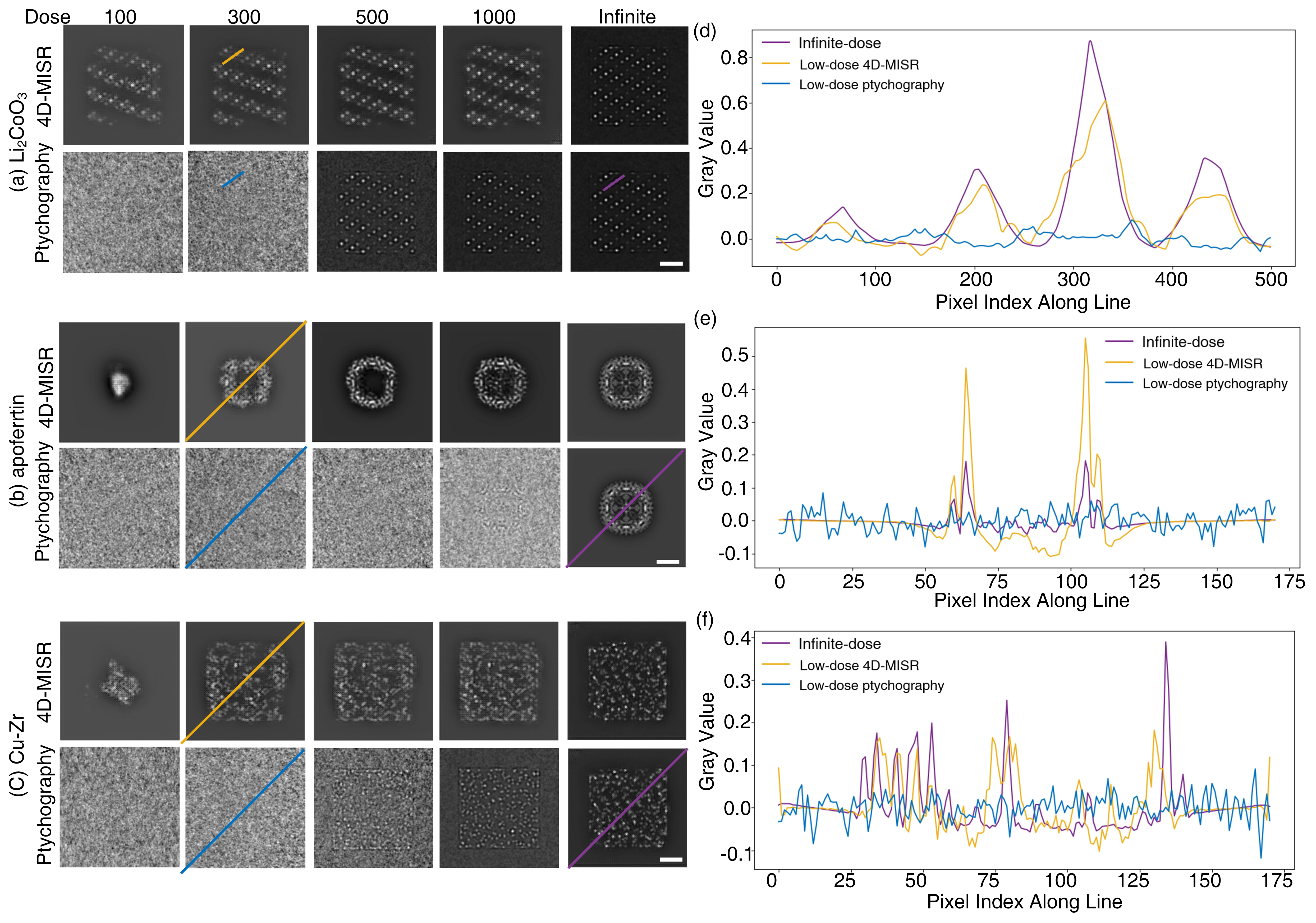} 
    \caption{\textbf{Comparison of 4D-MISR and ptychography reconstructions for various samples under different electron dose conditions.} 
    \justifying
    (a) Li$_2$CoO$_3$, (b) Cu-Zr metallic glass, and (c) apoferritin reconstructions are shown for electron doses ranging from 100 to 1000 e$^-$/\AA$^2$. For each sample, the top row presents images reconstructed using the 4D-MISR method, while the bottom row displays the corresponding reconstructions using ptychography. Qualitative comparison of infinite-dose ptychography, low-dose 4D-MISR, and low-dose ptychography for three representative samples: (d) Li$_2$CoO$_3$, (e) apoferritin, and (f) Cu-Zr metallic glass. Insets in the lower right corners of the highest-dose images indicate scale bars of 5 \AA{} (for panels a and b) and 50 \AA{} (for panel c). The electron doses used for the low-dose reconstructions are 300 e$^-$/\AA$^2$ for (d), (f) and (e).}
    \label{fig:4}
\end{figure}

To further evaluate the performance of 4D-MISR, we conducted tests on Li$_2$CoO$_3$ (2 nm thickness, Materials Project ID: mp-1173879), apoferritin (16.1 nm thickness, PDB ID: 8RQB)~\cite{kuccukouglu2024low} and Cu-Zr metallic glass (2 nm thickness), representing highly ordered, partially ordered and disordered structures, respectively, with varying sample sizes and scan step sizes. The resulting reconstructions were benchmarked against those obtained using the latest ptychographic reconstruction algorithms available in the \texttt{py4DSTEM} package, as shown in Figure~\ref{fig:4}. For each sample, images were reconstructed at increasing dose levels, allowing for a visual assessment of dose-dependent fidelity.

Across all three systems, 4D-MISR demonstrates markedly improved structural recovery at lower doses compared to ptychography. In the case of Li$_2$CoO$_3$, the periodic atomic arrangement is increasingly resolved with higher dose, but 4D-MISR begins to reveal lattice details even at 200~e$^{-}$/\AA$^{2}$, while ptychography fails to resolve features below 500~e$^{-}$/\AA$^{2}$.  In the apoferritin dataset, 4D-MISR resolves the hollow shell structure with increasing contrast as dose increases, whereas ptychographic reconstructions remain dominated by dose until the highest dose of 1000~e$^{-}$/\AA$^{2}$. A similar trend is observed in Cu-Zr, where 4D-MISR progressively reconstructs internal structural features, outperforming ptychography in both clarity and spatial coherence. 

The superior low-dose performance of 4D-MISR is largely due to its feature fusion framework, which reconstructs images by integrating features extracted from virtual bright-field (VBF) images synthesized from the 4D-STEM dataset. Under low-dose imaging conditions, convergent beam electron diffraction (CBED) suffers from greatly reduced electron counts per probe position, resulting in poor signal-to-noise ratios. This effect masks high-order diffraction details, thereby concealing high-spatial-frequency structural information beneath shot noise~\cite{li2025atomically}. Additionally, efforts to suppress background noise by decreasing the convergence semi-angle and detector collection angles reduce the accessible spatial-frequency bandwidth, fundamentally limiting resolution~\cite{kalinin2022machine}. This explains why traditional approaches like ptychography cannot reach atomic resolution under such constraints. The same degradation is observed in VBF images, as demonstrated in Supplementary Figure~1.


The resolution disparity becomes most significant under low-dose imaging conditions , highlighting 4D-MISR's enhanced resilience in dose-constrained regimes. This comparative analysis reveals that while both electron dose (100-infinite e$^{-}$/\AA$^{2}$) and sample thickness (2-20 nm) critically influence reconstruction fidelity, 4D-MISR achieves equivalent structural recovery with 60\% fewer incident electrons compared to conventional ptychography, which is a decisive advantage for radiation-vulnerable systems requiring strict dose budgets. Notably, our thickness-dependent studies demonstrate 4D-MISR's superior tolerance to dose variations across amorphous-crystalline interfaces. For relatively thin specimens such as the 2 nm Li$_2$CoO$_3$ sample shown in Fig.~\ref{fig:4}a, As the sample thickness increases, as seen in the 16.1 nm apoferritin in Fig.~\ref{fig:4}b, the image quality of both methods degrades due to increased multiple scattering and phase mixing. However, 4D-MISR demonstrates a higher tolerance to such thickness-induced artifacts, maintaining clearer structural features and better contrast under low-dose conditions. Nonetheless, for samples with thicknesses exceeding ~30 nm, the complexity of electron scattering may necessitate integration with three-dimensional reconstruction techniques such as multi-slice electron tomography to preserve reconstruction accuracy and reliability \cite{lee2023multislice}. These findings collectively establish an empirical framework for optimizing acquisition parameters in beam-sensitive material studies. 4D-MISR provides high-fidelity reconstructions that closely resemble the infinite-dose ground truth, significantly outperforming low-dose ptychography in terms of image clarity and structural detail.

We noted that 4D-MISR performs best at the center of the sample. Although the method does not reconstruct the full structure at 300~e$^{-}$/\AA$^{2}$ in Cu-Zr glass and apoferrtin, it produces high-quality results in the central region of the field of view. This is attributed to the multi-view geometry: multi-view imaging acquires information from multiple perspectives, resulting in overlapping coverage that is densest at the center. Peripheral regions, in contrast, often suffer from partial observability or self-occlusion. Prior work has shown that in multi-view reconstructions (e.g., three-view fusion), central regions maintain consistently high fidelity, whereas single-view approaches degrade near image margins or depth limits~\cite{wu2021multiview}. Geometric disparity under perspective projection further favors central regions: points near the optical axis undergo minimal parallax shifts across views, enabling more accurate alignment, while peripheral points exhibit greater projection distortion. As a result, the center benefits from more coherent feature fusion, yielding higher reconstruction quality.

To complement these observations on spatial resolution and region-specific performance, we next investigate two key imaging attribute: contrast and resolution. Contrast governs the perceptibility of such features, particularly under stringent low-dose conditions. Resolution determines the finest details that can be spatially distinguished. Therefore, a dedicated analysis of contrast characteristics offers additional insights into the robustness and practical utility of 4D-MISR in recovering structurally informative signals from dose-limited datasets\cite{clement2025ai}\cite{friedrich2023phase}.

To further evaluate the performance of our 4D-MISR method, we performed a contrast analysis based on line scan profiles extracted from the reconstructed images. Figure~\ref{fig:4} shows a comparison of contrast performance for three representative samples: crystalline Li$_2$CoO$_3$ (Fig.\ref{fig:4}a), biological apoferritin (Fig.\ref{fig:4}b) and Cu-Zr metallic glass (Fig.\ref{fig:4}c). For each sample, we compared three reconstructions: ptychography under low-dose conditions, our proposed 4D-MISR method, and an idealized infinite-dose ptychographic reconstruction serving as a reference. Line scans were drawn across representative regions in each reconstruction (horizontal for Li$_2$CoO$_3$ and diagonal for Cu-Zr metallic glass and apoferritin), as indicated by the red lines. The resulting intensity profiles are plotted alongside the images. These profiles quantify the gray-level variations along the scan direction and reflect the local contrast. In the Li$_2$CoO$_3$ sample, the line scan profile shows that 4D-MISR significantly enhances peak intensities compared to low-dose ptychography, closely approaching the profile of the infinite-dose reconstruction. The increased peak-to-background ratio in 4D-MISR indicates better contrast preservation of atomic columns, enabling more reliable structural interpretation. For the apoferritin sample, which is more sensitive to radiation damage, the diagonal line scan profile demonstrates that 4D-MISR produces substantially higher peak amplitudes compared to low-dose ptychography. This improvement suggests that our model effectively recovers high-contrast features even under challenging biological imaging conditions. Although it does not fully reach the contrast level of the infinite-dose reference, the 4D-MISR reconstruction bridges a significant portion of the gap. These results highlight the ability of our model to enhance contrast while maintaining low-dose acquisition, offering advantages for both materials and biological imaging applications in 4D-STEM. For the Cu-Zr metallic glass, 4D-MISR shows great potential in identifying the structures that contain valuable information. It is capable of accurately distinguishing the sample area from the non-sample area, and simultaneously performing focused imaging of the areas with high atomic density.

To quantitatively evaluate and compare the resolution performance between ptychography and our proposed 4D-MISR method, we conducted a frequency domain analysis based on the Fourier transforms of the reconstructed images. Specifically, we analyzed the isotropic power spectra of the reconstructions to determine the maximum resolvable spatial frequencies, corresponding to the edge of the detectable signal in the Fourier domain as shown in Supplementary Figure 3. This resolution improvement can be attributed to the capability of the 4D-MISR model to leverage multi-view contextual information and learn hierarchical frequency features from 4D-STEM data, thereby recovering more high-frequency components that are otherwise lost or attenuated in traditional phase retrieval-based reconstructions.

\subsection{Methods Comparison} 
\begin{figure}[htbp]
    \centering
    \includegraphics[width=0.95\textwidth]{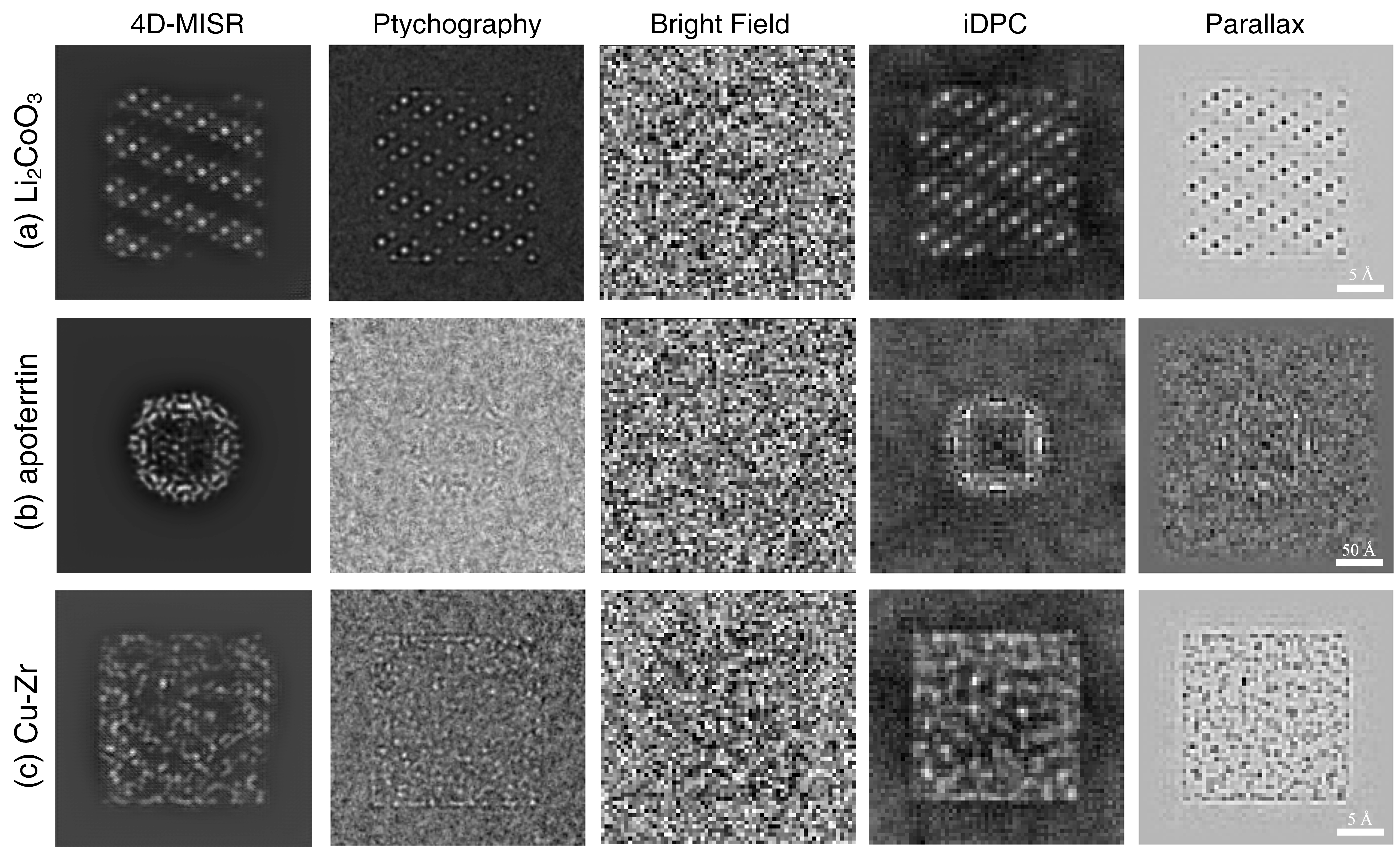} 
    \caption{\justifying
    Qualitative comparison of 4D-MISR with ptychography, bright-field imaging, integrated differential phase contrast (iDPC), and Parallax methods for three representative samples: (a) Li$_2$CoO$_3$, (b) Cu-Zr metallic glass, and (c) apoferritin. All reconstructions are performed using low-dose 4D-STEM datasets at 1000 e$^-$/\AA$^2$. Scale bars are 5 \AA{} for (a) and (b), and 50 \AA{} for (c).}
    \label{fig:5}
\end{figure}

We also compared virtual bright-field (VBF) imaging~\cite{savitzky2021py4dstem}, integrated Differential Phase Contrast (iDPC)~\cite{bosch2016integrated}, Parallax reconstruction~\cite{varnavides2023iterative}, ptychography~\cite{rodenburg2004phase}, and our proposed 4D-MISR method to comprehensively assess their respective advantages and limitations under low-dose 4D-STEM conditions on Li$_2$CoO$_3$, apoferritin, and Cu-Zr metallic glass, as shown in Figure~\ref{fig:5}. This comparison is not only necessary to benchmark the performance of 4D-MISR but also to contextualize its capabilities relative to existing techniques. Notably, during dataset generation, the pixelated detector was configured in "limit" mode to suppress spurious high-frequency signals. The spatial resolution of the first three methods depends on real-space sampling $(R_x, R_y)$, whereas ptychography is based on reciprocal space $(Q_x, Q_y)$. In contrast, 4D-MISR also relies on real space but at an upsampled resolution of $(rR_x, rR_y)$, with $r=3$ in our experiments.

 Each method embodies a distinct principle: VBF, while simple and dose-efficient, offers low contrast and limited resolution; iDPC enhances visibility of light elements through phase sensitivity but is constrained by detector segmentation and struggles with complex or thick samples; Parallax reconstruction introduces angular diversity to infer depth information and offers benefits in 3D imaging, yet suffers from acquisition complexity and resolution limitations; and ptychography, regarded as a gold standard for phase retrieval, provides high resolution through overlapping diffraction patterns, but its effectiveness plummets in low-dose regimes due to shot noise and the computational demands of phase reconstruction.

In contrast, 4D-MISR introduces a fundamentally different approach by leveraging multi-view virtual bright-field images within a dual-branch attention-guided deep learning framework. This allows it to adaptively emphasize high-SNR regions and intelligently integrate angular and spatial information, achieving superior reconstruction fidelity even at doses as low as 200 e$^-$/\AA$^2$. Especially in central regions of the imaging field, where multi-view overlap is greatest, 4D-MISR delivers sharper, higher-contrast images compared to traditional methods, which often fail to extract coherent details under similar dose constraints. Rather than relying solely on model-based Fourier inversion as in ptychography, 4D-MISR reinterprets the reconstruction problem as one of data-driven super-resolution, effectively recovering latent structural information that would otherwise be lost. Taken together, these results demonstrate that 4D-MISR is not only more robust under practical low-dose conditions but also scalable and compatible with existing 4D-STEM acquisition setups. Its unique capability to integrate angular redundancy through learning-based feature fusion makes it a promising and forward-looking solution for imaging radiation-sensitive materials and biological specimens where dose budgets are extremely limited.

\section{Conclusion}

In this study, we present 4D-MISR, an adventurous framework for high-resolution imaging of beam-sensitive materials through multi-view fusion of 4D-STEM datasets. Systematic comparisons with conventional ptychography and state-of-the-art deep learning methods reveal 4D-MISR’s unique capability to overcome critical limitations in low-dose regimes: while ptychography experiences from signal-to-noise constraints and computationally expensive phase retrieval, our unified framework achieves remarkable enhancement in contrast-to-noise ratio at doses below $1 \times 10^3 \mathrm{e}^-/\text{\AA}^2$. This breakthrough stems from integrating multi-view virtual bright-field (VBF) image fusion with attention-guided neural networks, enabling precise recovery of structural fingerprints.

Noted that central to 4D-MISR’s success is its dual-branch architecture, which synergizes spatial and angular features through attention-gated prioritization of critical structural signals. By leveraging geometric redundancies in 4D-STEM scans, especially within central imaging plane regions—the method resolves sub-Ångström features without requiring hardware modifications or specialized detectors. Therefore, the framework retains full compatibility with standard 4D-STEM workflows, offering a practical solution for radiation-sensitive systems.

\section*{Data availability}
The exact codebase for this work is unavailable to the public due to proprietary reasons. The 4D-MISR code is available on Github \href{https://github.com/ZifeiWang0313/4D-MISR}{https://github.com/ZifeiWang0313/4D-MISR}.

\section*{Acknowledgments}
This research is funded by the National Natural Science Foundation of China under Grant No. 52127801. The authors gratefully acknowledge this financial support, which made the research and the completion of this study possible.

\section*{Conflict of Interest}
The authors declare that they have no known competing financial interests or personal relationships that could have appeared to influence the work reported in this paper.

\section*{Author's contribution}
Zifei Wang, Yujun Xie conceived and led the project. Zifei Wang developed the model, performed simulations, and drafted the manuscript. Zian Mao co-supervised the study and contributed to model refinement and theoretical analysis. Xiaoya He implemented the algorithm and conducted training and visualization. Xi Huang and Haoran Zhang supported dataset preprocessing and validation. Chun Cheng and Shufen Chu assisted with simulation and data augmentation. Zhengting Hou provided materials science expertise and sample analysis. Xiaoqin Zeng and Yujun Xie supervised the research. All authors reviewed and approved the final manuscript.

\section*{References}
\bibliographystyle{iopart-num}
\bibliography{cas-refs}

\end{document}


\title{4D-HFN: Low-dose super-resolution 4DSTEM imaging with virtual BF via feature fusion}

\begin{table}[htbp]
\centering
\begin{tabularx}{0.8\linewidth}{>{\raggedright\arraybackslash}X >{\raggedright\arraybackslash}X}
\toprule
\textbf{Description} & \textbf{Value} \\
\midrule
Acceleration voltage &  300 kV \\
Step size ($\vec{r}$) & 0.4 \AA , 4 \AA \\
Convergence angle & 10 mrad \\
Defocus & 1500 \AA \\
Orientation & ${\in \{\left[0\ 0\ 1\right], \left[1\ 0\ 0\right], \left[0\ 1\ 0\right]\}}$ \\
Thickness & $<$ 30 \AA \\
Dose & $ 10^2 \dots10^5$ e$^-$/\AA$^2$ \\
Structures & 700 \\
\bottomrule
\end{tabularx}
\caption{Training dataset parameters used in 4D-MISR simulations, including electron beam settings, sample geometries, and structural diversity.}
\label{supple_table1}
\end{table}

\begin{table}[htbp]
\centering
\begin{tabularx}{0.8\linewidth}{>{\raggedright\arraybackslash}X >{\raggedright\arraybackslash}X}
\toprule
\textbf{Description} & \textbf{Value} \\
\midrule
Learning rate &  $ 0.5 \times 10^{-4}$ \\
Weight decay  & $ 1 \times 10^{-4}$ \\
Batch size & 32 \\
Batch count per feedback & 1 \\
Epoch & 512 \\
Seed & 1919810  \\
Optimizer & AdamW \\
\bottomrule
\end{tabularx}
\caption{Hyperparameters for training the 4D-MISR model, optimized to balance convergence stability and reconstruction quality.}
\label{supple_table2}
\end{table}
\begin{figure}[htbp]
    \centering
    \includegraphics[width=0.95\textwidth]{supplemental materials/S Fig1.jpg} 
    \caption{\textbf{4D-HFN reconstruction results across different dose conditions.} Single virtual bright field (VBF) images and corresponding 4D-HFN super-resolved reconstructions are shown under increasing dose levels. The results illustrate the network's ability to progressively recover structural features and contrast in both crystalline and amorphous materials even under stringent dose constraints.}
    \label{fig:s1}
\end{figure}


\begin{figure}[htbp]
    \centering
    \includegraphics[width=0.95\textwidth]{supplemental materials/S Fig3.jpg} 
    \caption{\textbf{Comparison of imaging performance between 4D-MISR and ptychography across different sample types.} (a) Li$_2$CoO$_3$, (b) Cu-Zr metallic glass, and (c) apoferritin protein complex. Left column: real-space reconstructions; right column: Fourier transforms showing maximum spatial frequency (red arrows). 4D-MISR consistently recovers higher resolution features than ptychography, notably achieving 3.3 \AA~vs. 4.7 \AA~in Li$_2$CoO$_3$. These results demonstrate 4D-MISR's applicability to diverse material systems.}
    \label{fig:s3}
\end{figure}